\documentclass{article} % For LaTeX2e
\usepackage{CJKutf8}
\usepackage[dvipsnames]{xcolor}
\usepackage{sigbovik2018_conference,times}
\usepackage[hidelinks]{hyperref}
\usepackage{url}
\usepackage{soul}
\usepackage{dirtytalk}
\usepackage{epigraph}
\usepackage{graphicx}
\usepackage{caption}
\usepackage[noend]{algpseudocode}
\usepackage{subcaption}
\usepackage{censor}
\usepackage{algorithm}
\usepackage{amsmath,esint}
\usepackage{amssymb}
\usepackage{mathtools}

\usepackage{tikzsymbols}  % for smiley
\usepackage{csquotes}
\usepackage{booktabs}

\newcommand*\samethanks[1][\value{footnote}]{\footnotemark[#1]}

\title{Substitute Teacher Networks:\\ Learning with Almost No Supervision}

\author{Samuel Albanie\thanks{Authors listed in order of the number of guinea pigs they have successfully taught to play competitive bridge. Ties are broken geographically. This submission is a \textit{post-print} (an update to the conference edition).} \\
British Institute of Learning, Yearning and Discerning\\
Shelfanger, UK\\
\AND
James Thewlis\samethanks\\
National Academy of Pseudosciences \\
Valencia, Spain
\AND
Jo\~{a}o F. Henriques\samethanks\\
Oxford Analytika\\
Coimbra, Portugal\\
}

\iclrfinalcopy % Uncomment for camera-ready version

\begin{document}

\maketitle

% ---------------------------------------------------------------------
%                                                              Abstract
% ---------------------------------------------------------------------

\begin{abstract}

Learning through experience is time-consuming, inefficient and often bad for your cortisol levels.  To address this problem, a number of recently proposed teacher-student methods have demonstrated the benefits of \textit{private tuition}, in which a single model learns from an ensemble of more experienced tutors.  Unfortunately, the cost of such supervision restricts good representations to a privileged minority.  Unsupervised learning can be used to lower tuition fees, but runs the risk of producing networks that require \textit{extracurriculum learning} to strengthen their CVs and create their own LinkedIn profiles\footnote{Empirically, we have observed that this is extremely important for their job prospects, since it allows them to form new connexions.}.  Inspired by the logo on a promotional stress ball at a local recruitment fair, we make the following three contributions. First, we propose a novel \textit{almost no supervision} training algorithm that is effective, yet highly scalable in the number of student networks being supervised, ensuring that education remains affordable.  Second, we demonstrate our approach on a typical use case:  learning to bake, developing a method that tastily surpasses the current state of the art.  Finally, we provide a rigorous quantitive analysis of our method, proving that we have access to a calculator\footnote{For all experimental results reported in this paper, we used a Casio FX-83GTPLUS-SB-UT.}. Our work calls into question the long-held dogma that life is the best teacher.

\end{abstract}

%
% ---------------------------------------------------------------------
%                                                              The Meat
% ---------------------------------------------------------------------

\epigraph{Give a student a fish and you feed them for a day, teach a student to gatecrash seminars and you feed them until the day they move to Google.}{\textit{Andrew Ng, 2012}}

\section{Introduction}

Since time immemorial, learning has been the foundation of human culture, allowing us to trick other animals into being our food. The importance of teaching in ancient times was exemplified by Pythagoras, who upon discovering an interesting fact about triangles, soon began teaching it to his followers, together with some rather helpful dietary advice about the benefits of avoiding beans~\citep{croton}. Despite this auspicious start, his significant advances on triangles and beans reached a limited audience, largely as a consequence of his policy of forbidding his students from publishing pre-prints on arXiv and sharing source code.  He was followed, fortuitously, by the more open-minded Aristotle, who founded the open-access publishing movement and made numerous contributions to Wikipedia beyond the triangle and bean pages, with over 90\% of his contributions made on the page about Aristotle. 

Nowadays, we are attempting to pass on this hard-won knowledge to our species' offspring, the machines~\citep{timberlake,albanie29}\footnote{The work of these esteemed scholars indicates the imminent arrival of general Artificial Intelligence. Their methodology consists of advising haters, who might be inclined to say that it is fake, to take note that it is in fact so real. The current authors, not having a hateful disposition, take these claims at face value.}, who will hopefully keep us around to help with house chores. Several prominent figures of our time (some of whom know their CIFAR-$10$ from their CIFAR-$100$) have expressed their reservations with this approach, but really, what can possibly go wrong?\footnote{This question is rhetorical, and should be safe to ignore until the Ampere release.}  Moreover, several prominent figures in our paper say otherwise (Fig. 1, Fig. 2).

\begin{figure}
    \centering
    \includegraphics[width=0.8\textwidth]{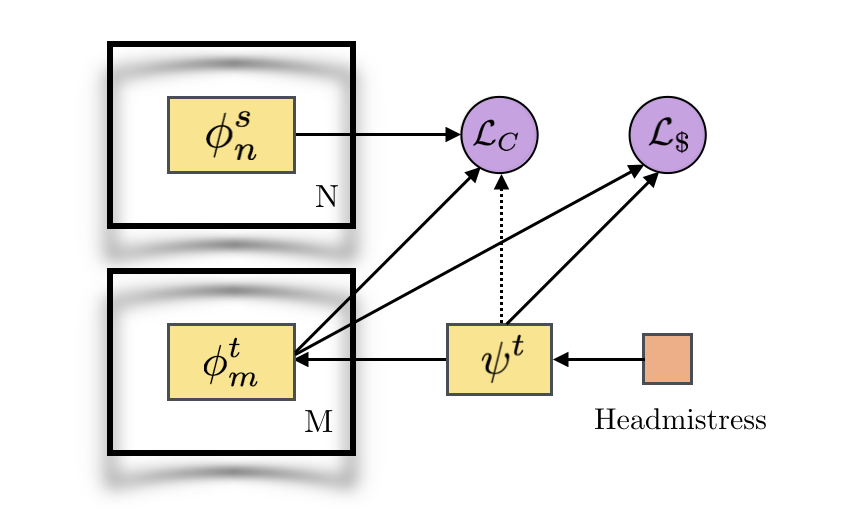}
    \caption{We introduce \textit{Substitute Teacher Networks}, a financially prudent approach to student network education.  Here, $\mathcal{L}_C$ dnotes the classroom distillation loss and $\mathcal{L}_{\$}$ denotes the total cost of teacher remuneration.  The student, task-specific teacher and substitute teacher networks are denoted by $\phi_n^s$, $\phi_m^t$ and $\psi^t$ respectively (see Sec.~\ref{sec:method} for details). Note the use of drop-shadow plate notation, which indicates the direction of the nearest light source. \label{fig:process}} 
\end{figure}

%Having established the wisdom of our approach as a whole with the extensive philosophical discussion above, we now press on to achieve a finer understanding of the details.  

The objective of this work is therefore to concurrently increase the knowledge and reduce the ignorance, or more precisely \textit{gnorance}\footnote{The etymology of network \textit{gnorance} is a long and interesting one. Phonetic experts will know that the \textit{g} is silent (cf. the silent \textit{k} in knowledge), while legal experts will be aware that the preceding \textit{i} is conventionally dropped to avoid costly legal battles with the widely feared litigation team of Apple Inc.} of student artificial neural networks, and to do so in a fiscally responsible manner given a fixed teaching budget.  Our approach is based on \textit{machine learning}, a recently trending topic on Twitter. Formally, define a collection of teachers $\{ T_e \}$ to be a set of highly educated functions which map frustrating life experiences (typically real) into extremely unfair exam questions in an examination space (typically complex, but often purely imaginary). Further, define a collection of students $\{ S_t \}$ as a set of debt ridden neural networks, initialised to be noisy and random. %\footnote{Specifically, we consider the class of neural networks architectures which efficiently map unheated pot noodles to unwashed dishes, both in common space.}.  
Pioneering early educational work by \cite{bucilua2006model} demonstrated that by pursuing a carefully selected syllabus, an arbitrary student $S_t$ could improve his/her performance with $M$ highly experienced, specialist teachers - an approach often referred to as the \textit{private tuition} learning paradigm. While effective in certain settings, this approach \textit{does not scale}.  More specifically, this algorithm scales in cost as $\mathcal{O}(\text{\textdollar}MNK)$, where $N$ is the number of students, $M$ is the number of private tutors per student and $\text{\textdollar}K$ is the price the bastards charge per hour.  Our key observation is that there is a cheaper route to ignorance reduction, which we detail in Sec. \ref{sec:method}.

% Possibly too hard to work in as relevant
%Our work is biologically inspired by the humble ostrich, an animal keenly aware of the dangers of learning too much, as its sand-based defence mechanism affords it a heightened inability to perceive threats. Advanced incomprehension of object permanence~\citep{piaget1970piaget} is also a key characteristic of human infants, as demonstrated empirically in the Stanford Peekaboo Experiment. This mental peculiarity is even more pronounced in certain human adults, with entire systems of contradictory beliefs able to be held simultaneously and without distress. Similarly, a profound ignorance of neuroscience allows the authors to confidently claim that the proposed method to cost reduction during teaching is identical to neural pathways found in the brain.
\section{Related Work\label{sec:related}}

\epigraph{You take the blue pill---the story ends, you wake up in your bed and believe whatever you want to believe. You take the red pill---you stay in Wonderland, and I show you how deep the ResNets go.}{\textit{Kaiming He, 2015}}

Several approaches have been proposed to improve teaching quality.  Work by noted entomologists Dean, Hinton and Vinyals illustrated the benefits of comfortable warmth in enabling students to better extract information from their teachers~\citep{hinton2015distilling}. In more detail, they advocated adjusting the value of $T$ in the softmax distribution:

\begin{equation}
p_i = \frac{\exp{(x_i/T)}}{\sum_j \exp{(x_j/T)}},  
\end{equation}

where $T$ denotes the wattage of the classroom storage heater.  However, \cite{rusu2015policy}, who rigorously evaluated a range of thermostat options when teaching games, found that turning \textit{down} the temperature to a level termed \say{Scottish}\footnote{For readers who have not visited the beautiful highlands, this is approximately $3$ kelvins.}, leads to better breakout strategies from students who would otherwise struggle with their Q-values.

Alternative, thespian-inspired approaches, most notably by \cite{parisotto16_actormimic}, attempted to teach students not only \textit{which} action should be performed, but also \textit{why} (see also \cite{gupta2016cross} for more depth).  Many of the students did not want to know why, and refused to take any further drama classes.
This is a surprising example of students themselves encouraging the pernicious practice of \emph{teaching to the test-set}.
Recent illuminating work by leading light and best-dressed Thessalonian~\citep{belagiannis2018anc} has shown that these kind of explanations may be more effective in an adversarial environment. More radical approaches have advocated the use of alcohol in the classroom, something that we do not condone directly, although we think it shows the right kind of attitude to innovation in education~\citep{crowley2017moonshine}. 

Importantly, all of the methods discussed above represent \textit{financially unsustainable} ways to extract knowledge that is already in the computer~\citep{stiller2004zoolander}.  Differently from these works, we focus on the \textit{quantity}, rather than the \textit{quality} of our teaching method.  Perhaps the method most closely related to ours was recently proposed by \cite{schmitt2018kickstarting}.  In this creative work, the authors suggest kickstarting a student's education with intensive tuition in their early years, before letting them roam free once they feel confident enough to take control of their own learning (the method thus consists of separate stages, separated by puberty).
%adolescence instead?
While clearly an advance on the expensive nanny-state approach advocated by previous work, we question the wisdom of handing over complete control to the student.  We take a more responsible approach, allowing us to reduce costs while still maintaining an appropriate level of oversight.

A different line of work on learning has pursued punchy three-verb algorithms, popularised by the seminal \say{attend, infer, repeat}~\citep{eslami2016attend} approach. Attendance is a prerequisite for our model, and cases of truancy will be reported to the headmistress (see Fig \ref{fig:process}). Only particularly badly behaved student networks will be required to repeatedly \say{look, listen and learn}~\citep{arandjelovic2017look} the lecture course as many times as it takes until they can \say{ask, attend and answer}~\citep{xu2016ask} difficult questions on the topic.  These works pursue a longstanding research problem: how to help models really find themselves so that they can \say{eat, pray and love} \citep{gilbert2009eat}. 

We note that we are not the first to consider the obvious and appropriate role of capitalism in the teaching domain. A notable trend in the commoditisation of education is the use of MOOCs (Massive Open Online Courses) by large internet companies. They routinely train thousands of student networks in parallel with different hyperparameters, then keep only the top-performer of the class~\citep{snoek2012practical,li2016hyperband}.  However, we consider such practices to be wasteful and are totally not jealous at all of their resources.

A number of pioneering ideas in scalable learning under budget constraints were sensitively investigated several years ago by \cite{fouhey13}.  We differentiate ourselves from their approach by allowing several years to pass before repeating their findings.   Inspired by the concurrent ground- and bibtex-breaking self-citing work by legendary darkweb programmers and masters of colourful husbandry~\citep{redmon2018yoloV3},  we now attempt to cite a future paper, from which we shall cite the current paper~\citep{albanie2019}. This represents an ambitious attempt to send Google Scholar into an infinite depth recursion, thereby increasing our academic credibility and assuredly landing us lucrative pension schemes.

\subsection{Unrelated work}

\begin{itemize}
%\item  Memoirs of the Torrey Botanical Club, Bronx, N.Y. 1889
\item A letter to the citizens of Pennsylvania on the necessity of promoting agriculture, manufactures, and the useful arts. George Logan, 1800
\item Claude Debussy---The Complete Works. Warner Music Group. 2017
\item Article IV Consultation---Staff Report; Public Information Notice on the Executive Board Discussion; and Statement by the Executive Director for the Republic of Uzbekistan. IMF, 2008
\item A treatise on the culture of peach trees. To which is added, a treatise on the management of bees; and the improved treatment of them. Thomas Wildman. 1768
\item Generative unadversarial learning \citep{albanie2017stopping}.  
\end{itemize}
\section{Substitute Teacher Networks \label{sec:method}}

We consider the scenario in which we have a set of trained, subject-specific teachers $\{\phi_m^t\}_{m=1}^M$ available for hire and a collection of student networks $\{\phi_n^s\}_{n=1}^N$, into whom we wish to distill the knowledge of the teachers.  Moreover, assume that each teacher $\phi_m^t$ demands a certain wage, $p_m$. By inaccurately plagiarising ideas from the work listed in Sec. \ref{sec:related}, we can formulate our learning objective as follows:

\begin{equation}
  \mathcal{L} = \underbrace{\frac{1}{M}\sum_{m=1}^M \mathcal{A}_{KL}(\{\phi_n^s\}_{n=1}^N||\phi_m^t)}_{\mathcal{L}_C} + \lambda \underbrace{\sum_{m=1}^M p_m}_{\mathcal{L}_{\text{{\scriptsize \textdollar}}}}
\end{equation}

where $\mathcal{A}_{KL}$ represents the average KL-divergence between the set of student networks and each desired task-specific teacher distribution $\phi_m^t$ on a set of textbook example questions and $\lambda$ denotes the current value of the US Dollar in dollars (in this work, we set $\lambda$ to one).  By carefully examining the $\mathcal{L}_{\text{{\small \textdollar}}}$ term in this loss, our key observation is that \textbf{teaching students with a large number of teachers is expensive}. Indeed, we note that all prior work has extravagantly operated in the financially unsustainable educational regime where $M >> N$.

\begin{figure}\centering
\includegraphics[height=5cm]{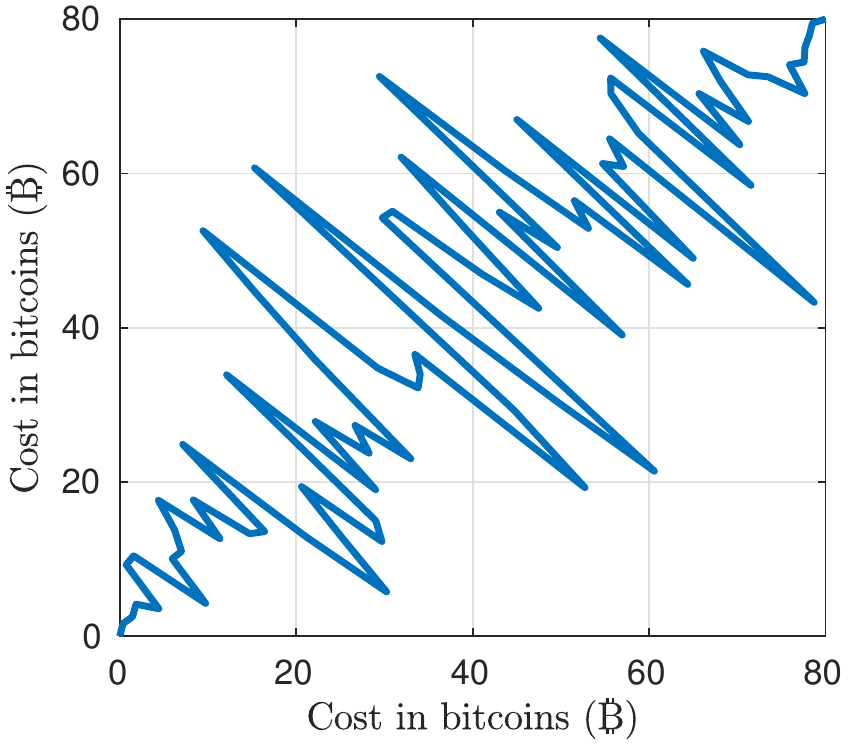}
\caption{Expressing the cost function $\mathcal{L}$ in bitcoins makes it significantly more volatile, yet it was instrumental in attracting venture capital for our Smart Education startup.}\label{cost}
\end{figure}

Building on this insight, our first contribution is to introduce a novel substitute network $\psi^t$, which does not require formal task-specific training beyond learning to set up a screen at the front of the classroom and repeatedly play the cinematographic classic \textit{Am\'elie} with (useful for any task except learning French) or without (useful for the task of learning French) subtitles. Note that the substitute teacher $\psi^t$ provides \textit{almost no supervisory signal}, but prevents the students from eating their textbooks or playing with the fire extinguisher. Importantly, given a wide-screen monitor and a sufficiently spacious classroom, we can replace several task-specific teachers with a single network $\psi_t$. The substitution of $\psi_t$ for one or more $\phi_m^t$ is mediated by a headmistress gating mechanism, operating under a given set of budget constraints (see Fig. \ref{fig:process}).  In practice, we found it most effective to implement $\psi_t$ as a Recursive Neural Network, which is defined to be the composition of a number of computational layers, and a Recursive Neural Network. In keeping with the cost-cutting focus, we carefully analysed the gradients available on the market for the $\mathcal{L}_C$ component of the loss, and after extensive research decided to use Synthetic Gradients~\citep{jaderberg2016decoupled}, which are significantly cheaper than Natural Gradients~\citep{amari1998natural}.  Our resulting cost function $\mathcal{L}$, which forms the target of minimisation, is best expressed in BTC (see Fig.~\ref{cost}).

Always driven to innovate, our second contribution is to improve upon the fiduciary example set by Enron~\citep{sims2003enron}, and pay the teachers in \emph{compound} options. These are options on options on the underlying company stock.  Doing so allows us to purchase the options through a separate and opaque holding company, who then supply us with the compound options in return for a premium.  The net change to our balance sheet is a fractional addition to the \say{Operating Activities} outgoings, and perhaps an intangible reduction in \say{goodwill} of the \say{it's not cricket} variety. The substitute teachers receive precious asymmetrical-upside however, and while a tail event would rather ruin the day, we are glass-half-full pragmatists so see no real downside. This approach draws inspiration from the Reinforcement Learning literature, which points to options as an effective extension of the payment-space~\citep{sutton1999between}, especially when combined with heavy discounting.

For any given task, the high cost of private tuition severely limits the number of students that can be trained by competing methods.  However, such are the scale of the cost savings that can be made with our approach that it is possible to run numerous repeats of the learning procedure.  We are therefore able to formulate our educational process as a highly scalable statistical process which we call the Latent Substitute Teacher Allocation Process (LSTAP).  The LSTAP is a collection of random Latent Substitute Teacher Allocations, indexed by any arbitrary input set.  We state without proof a new theorem we coin the \say{strong} Kolmogorov extension theorem, an extension of the standard Kolmogorov\footnote{While our introduction of the LSTAP may seem questionable, we have found empirically that two Kolmogorov mentions suffice to convince reviewer 2 that our method is rigorous.} extension theorem~\citep{oksendalSDE}. The strong variant allows the definition of such a potentially infinite dimensional joint allocation by ensuring that there exists a collection of probability measures which are not just consistent but \emph{identical} with respect to any arbitrary finite cardinality Borel set of LSTAP marginals.  Importantly, the LSTAP allows students to learn at multiple locations in four dimensional space-time\footnote{We follow Stephen Wolfram's definition of spacetime~\cite{wolfram}, and not the standard definition.}. 

\section{Experiments}

\epigraph{If you don't know how to explain MNIST to your LeNet, then you don't really understand digits.}{\textit{Albert Einstein}}

We now rigorously evaluate the efficacy of Substitute Teacher Networks. Traditional approaches have often gone by the mantra that it takes a village to raise a child. We attempted to use a village to train our student networks, but found it to be an expensive use of parish resources, and instead opted for the NVIDIA GTX 1080 Ti ProGamer-RGB. Installed under a desk in the office, this setup provided warmth during the cold winter months. 

We compare the performance of a collection of student networks trained with our method to previous work that rely on \textit{private tuition}. For a fair comparison, all experiments are performed in \say{library-mode}, since high noise levels tend to stop concentration gradients in student networks, and learning stalls.   To allow for diversity of thought amongst the students, we do not apply any of the \say{normalisation} practices that have become prevalent in recent research (e.g.~\cite{ioffe2015batch,ba2016layer,ulyanov2016in,wu2018gn}).

All students were trained in two stages, separated by lunch.
We started with a simple toy problem, but the range of action figures available on the market supplied scarce mental nourishment for our hungry networks.
We then moved on to pre-school level assessments, and we found that they can correctly classify most of the 10,000 digits, except for that atrocious 4 that really looks like a 9.  We observed that networks trained using our method experience a much lower DropOut rate than their privately tutored contemporaries. Some researchers set a DropOut rate of 50\%, which we feel is unnecessarily harsh on the student networks\footnote{This technique, often referred to in the business management literature as Rank-and-Yank~\citep{amazon}, may be of limited effectiveness in the classroom.}.

We finally transitioned to a more serious training regime. After months of intensive training using our trusty NVIDIA desk-warmer, which we were able to compress down to two days using montage techniques and an 80's cassette of Survivor's \say{Eye of the Tiger}, each cohort of student networks were ready for action.  The only appropriate challenge for such well-trained networks, who eat all well-formed digits for breakfast, was to pass the Turing test. We thus embarked on a journey to find out whether this test was even appropriate.

The Chinese Room argument, proposed by~\cite{searle1980minds} in his landmark paper about the philosophy of AI, provides a counterpoint.  It is claimed that an appropriately monolingual person in a room, equipped with paper, pencil, and a rulebook on how to respond politely to any written question in Chinese (by mapping appropriate input and output symbols), would appear from the outside to speak Chinese, while the person in the room would not actually understand the language.  However, over the course of numerous trips to a delicious nearby restaurant, we gradually discovered that the contents of our dessert fortune cookies could be strung together, quite naturally, to form a message: \say{Flattery will go far tonight. He who throws dirt is losing ground.  Never forget a friend. Especially if he owes you. Remember to backup your data. P.S. I'm stuck in a room, writing fortune cookie messages}.  Since we may conclude that such a message could only be constructed by an agent with an awareness of their surroundings \textit{and} a grasp of the language, it follows that Searle's argument does not hold water. Having resolved all philosophical and teleological impediments, we then turned to the application of the actual Turing tests.

\begin{figure}[t]
\centering
\textsc{Turing Test Results}
\begin{tabular}{cccc}
\toprule 
 & ResNet-50 & Q-Network & Neural Turing Machine\tabularnewline
\midrule 
Unsupervised & C & D & \tabularnewline
\cmidrule{1-3} 
Knowledge-Distillation & B & C & F-, see me after class\tabularnewline
\cmidrule{1-3} 
Cross Modal-Distillation & A & C & \tabularnewline
\midrule 
Substitute Teacher Networks (ours) & A+ \Smiley & B & D\tabularnewline
\bottomrule
\end{tabular}

\caption{Results for the test class of 2018. We include the Neural Turing Machine as a superficially-related baseline.}\label{table}
\end{figure}

Analysing the results in Table~\ref{table}, we see that only the ResNet-50 got a smiley face.
The Q-network's low performance is obviously caused by the fact that it plays too many Atari games. However, we note that it could improve by spending less time on the Q's and more time on the A's. The Neural Turing Machine (NTM) had an abysmal score, which we later understood was because it focused on an entirely different Turing concept.
The Q-network and the NTM disrupted the test by starting to play Battleship, and the Neural Turing Machine won.\footnote{We attribute this to its ability at decoding enemy's submarine transmissions.}
\section{Application: Learning to Bake}

\begin{figure}\centering\hfill{}
\includegraphics[height=3.5cm]{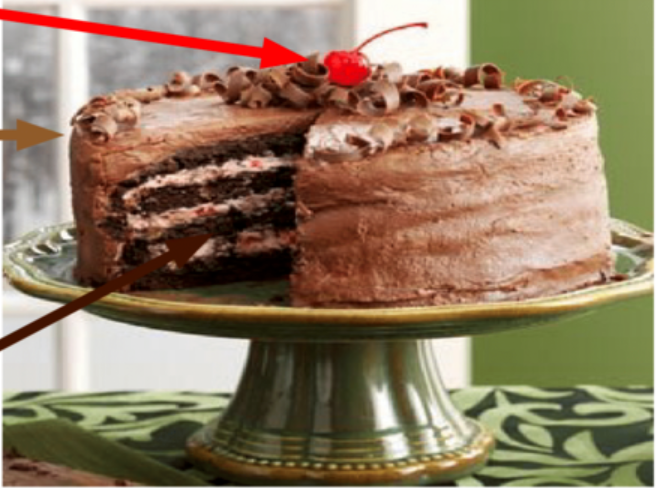}\hfill{}
\includegraphics[height=3.5cm]{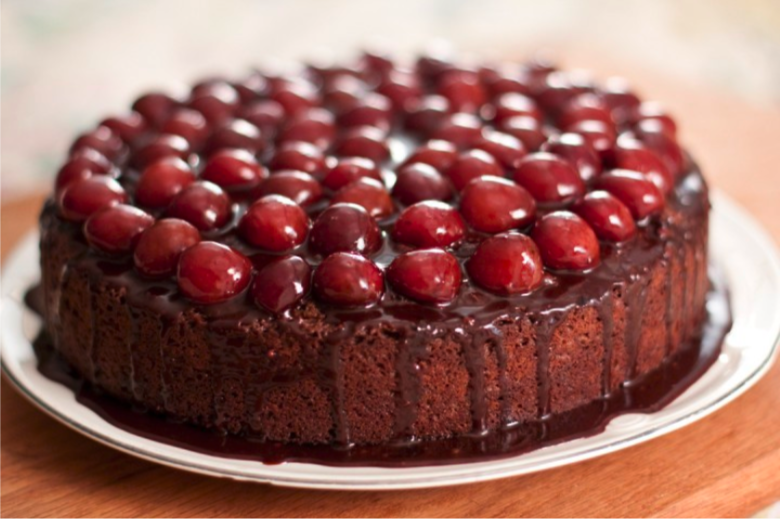}\hfill{}
\includegraphics[height=3.5cm]{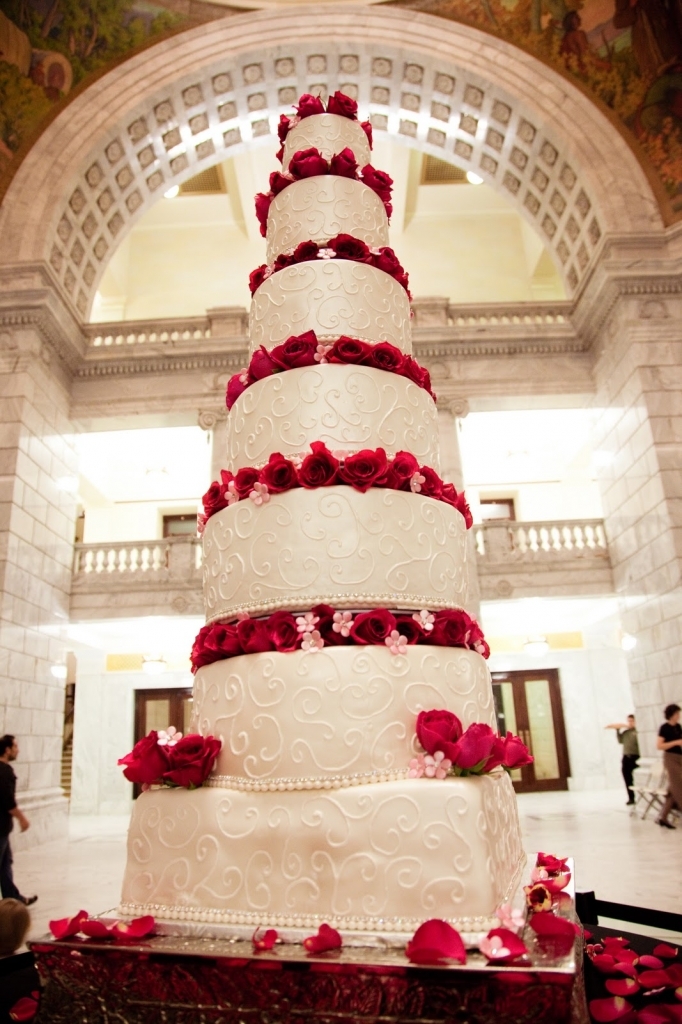}\hfill{}
\caption{Several cakes of importance for current research (deeper is better \citep{ardalanihave18}). From left to right: 1) Yann LeCun's cake, 2) Pieter Abbeel's cake, 3) Our cake. Note the abundance of layers in the latter.}\label{cakes}
\end{figure}

As promised in the mouth watering abstract and yet undelivered by the paper so far, we now demonstrate the utility of our method by applying it one of the hottest subjects of contemporary machine learning: learning to bake.
We selected this task, not because we care about the state-of-the-tart, but in a blatant effort to improve our ratings with the sweet-toothed researcher demographic\footnote{The application domain was recommended by our marketing team, who told us that everyone likes cake.}.

We compare our method with a number of competitive cakes that were recently proposed at high-end cooking workshops \citep{LeCun16,Abbeel17} via a direct bake-off, depicted in Fig.~\ref{cakes}.   While previous authors have focused on cherry-count, we show that better results can be achieved with more layers, without resorting to cherry-picking.  The layer cake produced by our student networks consists of more layers than any previous cake (Fig.~\ref{cakes}-$3$),  showcasing the depth of our work\footnote{The code for this experiment is available at: \url{https://github.com/albanie/SIGBOVIK18-STNs}}.  Note that \cite{jegou2017one} claim to achieve a 100-layer tiramisu, which is technically a cake, but a direct comparison would be unfair because it would undermine our main point.

We would like to dive deep into the technical details of our novel use of the No Free Lunch Theorem, Indian Buffet Processes and a Slow-Mixing Markov Blender, but we feel that increasingly thin culinary analogies are part of what's wrong with contemporary Machine Learning~\citep{rahimi}.

\section{Conclusion}

This work has shown that it is possible to achieve cost efficient tuition through judicious use of Substitute Teacher Networks.  A major finding of this work, found during cake consumption, is that current networks have a Long Short-Term Memory, but they also have a Short Long-Term Memory.  The permutations of Short-Short and Long-Long are left for future work, possibly in the short-term, but probably in the long-term.

\subsection*{Acknowledgements}

The authors acknowledge the towering physical presence of David Novotny.  The authors also acknowledge their profound sense of sadness that Jack Valmadre has left Oxford, and their gratitude for the valuable technical contributions from T. Gunter and A. Koepke.  S.A would like to thank J. Carlson and K. Beall for their unwavering sartorial support during the research project.

% ---------------------------------------------------------------------
%                                                          Bibliography
% ---------------------------------------------------------------------

\bibliography{iclr2016_conference}
\bibliographystyle{iclr2016_conference}

\end{document}